# Dirichlet Process with Mixed Random Measures: A Nonparametric Topic Model for Labeled Data

**Dongwoo Kim** DW.KIM@KAIST.AC.KR
Computer Science Department, KAIST, Daejeon, Korea

**Suin Kim** SUIN.KIM@KAIST.AC.KR
Computer Science Department, KAIST, Daejeon, Korea

**Alice Oh** ALICE.OH@KAIST.EDU
Computer Science Department, KAIST, Daejeon, Korea

## Abstract

We describe a nonparametric topic model for labeled data. The model uses a mixture of random measures (MRM) as a base distribution of the Dirichlet process (DP) of the HDP framework, so we call it the DP-MRM. To model labeled data, we define a DP distributed random measure for each label, and the resulting model generates an unbounded number of topics for each label. We apply DP-MRM on single-labeled and multi-labeled corpora of documents and compare the performance on label prediction with MedLDA, LDA-SVM, and Labeled-LDA. We further enhance the model by incorporating ddCRP and modeling multi-labeled images for image segmentation and object labeling, comparing the performance with nCuts and rddCRP.

## 1. Introduction

Topic models such as latent dirichlet allocation (LDA) (Blei et al., 2003) have been extended to incorporate side information such as authorship (Rosen-Zvi et al., 2004), spatial or temporal coordinates (Wang & Grimson, 2007; Wang et al., 2008), and document labels (Ramage et al., 2009). Most of these models are parametric topic models, and they cannot be simply converted to nonparametric counterparts which generally have various advantages over parametric models. In the Bayesian nonparametric (BNP) literature on Dirichlet processes (DP), modeling unknown densities with covariates has been often done with dependent Dirichlet Processes (DDP), but extending DDP for topic modeling requires more complex model settings and posterior inferences (Srebro & Roweis, 2005).

In this paper, we propose a novel nonparametric topic model, Dirichlet process with mixed random measures (DP-MRM) for documents with an arbitrary amount of discrete side information such as labels. DP-MRM can be seen as a nonparametric extension of Labeled-LDA (L-LDA) (Ramage et al., 2009) in terms of defining topic distributions over labels. Recent research shows that incorporating label information into topic models has advantages for topic interpretation as well as other practical uses such as user profiling in social media (Ramage et al., 2010). However, L-LDA assumes that each label corresponds to a single multinomial (i.e., topic), and a document is only generated by the topics of the observed labels. Consequently, the model imposes an overly limiting restriction on the topics with which to represent the documents. While L-LDA models each label with a single multinomial, DP-MRM models each label with a random measure which is defined over the entire topic space.

There are several supervised topic models including sLDA (Blei & McAuliffe, 2007), discLDA (Lacoste-Julien et al., 2008), and medLDA (Zhu et al., 2009), that also model data with labels. There are two major differences between those models and DP-MRM. First, in the former models which are designed specifically for classification, each label acts as the supervisor for learning. In L-LDA and DP-MRM which are designed with the focus on understanding the meaning of each label in terms of the latent topics, each label actually is the label for one (in L-LDA) or a set of (DP-MRM) topic(s). Second, the former models are restricted to modeling data with one label per document and cannot model documents with multiple labels. To illustrate this second point, we evaluate DP-MRM on data with single labels as well as multiple labels.

Another view of DP-MRM is that it is a more general case of the HDP (Teh et al., 2006). Modeling the corpus with our model using a single label for all documents would produce the same results as the HDP. Viewed this way, DP-MRM can be used instead of the HDP in many BNP models that are extensions of HDP. We show an example of this by incorporating the ddCRP (Blei & Frazier, 2011) into our model for the task of image segmentation as done in rddCRP (Ghosh et al., 2011).

The paper is organized as follows. In section 2, we describe DP-MRM along with the stick-breaking and Pólya urn perspectives. In section 3, we derive a sampling method for the latent variables based on Gibbs sampling. In section 4, we demonstrate our approach on labeled documents for single-labeled and multi-labeled corpora and compare the performance of our model by label prediction and heldout likelihood against LDA-SVM and L-LDA. In section 5, we present a modification of our model for image segmentation and compare the performance with nCuts (Shi & Malik, 2000) and rddCRP (Ghosh et al., 2011) quantitatively and qualitatively.

## 2. Dirichlet Process with Mixture of Random Measures

In this section, we describe our model, Dirichlet process with mixed random measures (DP-MRM) model. We first review the generative process of L-LDA, and then we show how DP-MRM incorporates label information within the BNP framework based on Dirichlet Processes (DP). Lastly, we present the stick breaking process and the Pólya urn scheme for DP-MRM.

### 2.1. Model Definition

L-LDA is a supervised version of LDA for modeling multi-labeled documents. The generative process of L-LDA starts with a definition of a document specific function $label(j)$, which returns a set of observed label indices for document $j$. Then, for each document $j$, a multinomial distribution $\theta_j$ over topics is randomly sampled from a Dirichlet with parameter $r_j \alpha$, where $r_j$ is a $K$ dimensional vector whose $k$th value is 1 if $k \in label(j)$ and 0 if $k \notin label(j)$. Then, to generate the word $i$, a topic $z_{ji}$ is chosen from this topic distribution, and a word, $x_{ji}$, is generated by randomly sampling from a topic-specific multinomial distribution $\phi_{z_{ji}}$. By using a document specific indicator vector $r_j$, the model can specify the topic proportion of document $\theta_j$ over the $|label(j)|-1$ dimension simplex.

We now describe the generative process of Dirichlet process with mixed random measures. First, we define a DP distributed random measure $G_0^1, ..., G_0^K$ over $K$ possible labels with a base distribution $H$ as follows:

$$H \mid \beta \equiv \text{Dir}(\beta)$$
$$G_0^k \mid \gamma_k, H \sim \text{DP}(\gamma_k, H), \quad (1)$$

where the base distribution $H$ is assumed to be a symmetric Dirichlet distribution over the entire vocabulary dimension, and $\gamma_k$ controls the variability of $G_0^k$. By defining one random measure per label, we place an infinite topic space for each label. For each document $j$, another DP distributed random measure $G_j$ is defined with a mixture of labeled-random measure as follows:

$$\lambda_j \sim \text{Dir}(r_j \eta)$$
$$G_j | label(\cdot), \alpha, \lambda_j \sim \text{DP}(\alpha, \sum_{\{k; label(j)\}} \lambda_{jk} G_0^k) \quad (2)$$

where $\alpha$ is a concentration parameter, $\lambda_{jk}$ is a mixing proportion of $G_0^k$, and $\eta$ controls the sparsity of $\lambda_j$. DP-MRM uses a mixture of random measures, $\sum_k \lambda_{jk} G_0^k$, as the base distribution of $G_j$, the document specific measure. For the mixing proportion $\lambda_{jk}$ of each $G_0^k$, we sample $\lambda_j$ from a symmetric Dirichlet prior parameterized by $r_j$ and $\eta$. Hence, with the observed labels $label(j)$, $r_j$ selectively specifies the mixing proportions of $G_0^k$ over the $|label(j)|-1$ dimensional simplex.

For each word $x_{ji}$ in document $j$, the probability of drawing a word $x_{ji}$ is parameterized by a random variable $\theta_{ji}$ drawn from $G_j$ with some family of distribution $F$. It is typically assumed to be a multinomial distribution,

$$\theta_{ji} \mid G_j \sim G_j$$
$$x_{ji} \mid \theta_{ji} \sim F(\theta_{ji}) \qquad F(\theta_{ji}) \equiv \text{Mult}(\theta_{ji}), \quad (3)$$

which makes $F$ to be conjugate to the base distribution $H$, and so it is possible to integrate out the factors $\theta_{ji}$.

As a result of the construction, the model chooses an appropriate number of topics for each label. Note that HDP can be viewed as a specialized instance of our model (Teh et al., 2006), where we assume there is a single 'unknown' label for all documents. Then the overall corpus is defined by a set of topics from the single 'unknown' label, $G_0^{\text{unknown}} \sim \text{DP}(\gamma, H)$, and the random measure for document $j$ is drawn from $G_j \sim \text{DP}(\alpha, G_0^{\text{unknown}})$. A similar idea of using a mixture of random measures was proposed in (Antoniak, 1974), but our model extends that idea into a hierarchical construction for the grouped clustering problem.

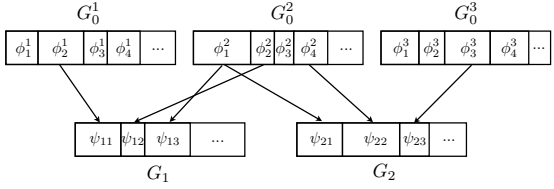

*Figure 1.* Example of sharing structure between first and second level DPs. $G_1$ is dirichlet distributed with the mixed base distribution $G_0^1$ and $G_0^2$ where $\phi_2^1 = \psi_{11}$, $\phi_2^2 = \psi_{12}$, and $\phi_1^2 = \psi_{13}$.

### 2.2. Construction and Predictive Distribution

We now describe two perspectives that are important for the inference algorithms for DP-MRM: the stick breaking process and the Pólya urn scheme.

**Stick breaking process** The stick breaking process is a constructive definition for generating a Dirichlet process (Sethuraman, 1991). Same as the model definition in the previous section, the stick breaking process can be divided into two level DPs. For the first level random measure $G_0^k$, we follow the general stick breaking process, which is given by the following conditional distributions:

$$v_l^k \sim \text{Beta}(1, \gamma_k) \qquad \pi_l^k = v_l^k \prod_{d=1}^{l-1}(1 - v_d^k)$$
$$\phi_l^k \sim H \qquad G_0^k = \sum_{l=0}^{\infty} \pi_l^k \delta_{\phi_l^k}, \qquad (4)$$

where $\delta$ is a Dirac delta measure. A general stick breaking process can be seen as two independent sequences of deciding the stick length $\pi_l$ by samples from i.i.d. Beta trials and deciding the atom of the $l$th stick $\phi_l$ by i.i.d. samples from $H$.

The second level stick breaking construction is given by the following conditional distributions:

$$\lambda_j \sim \text{Dir}(\boldsymbol{r}_j \eta)$$
$$w_{jt} \sim \text{Beta}(1, \alpha) \qquad \pi_{jt} = w_{jt} \prod_{d=1}^{t-1}(1 - w_{jd})$$
$$k_{jt} \sim \text{Mult}(\lambda_j) \qquad \psi_{jt} \sim G_0^{k_{jt}}$$
$$G_j = \sum_{t=0}^{\infty} \pi_{jt} \delta_{\psi_{jt}}. \qquad (5)$$

Deciding the length of each stick is the same as the general stick breaking process, but assigning atoms for each divided stick must be changed because there are $K$ random measures for drawing $\psi_{jt}$. We introduce $k_{jt}$ as an indicator to $G_0^k$ where atom $\psi_{jt}$ is drawn.

We let $\theta_{ji}$ denote the random variable drawn from $G_j$, $\psi_{jt}$ the atom of $G_j$, and $\phi_l^k$ the atom of $G_0^k$. Note that each $\theta_{ji}$ is associated with one $\psi_{jt}$ (i.e., $\theta_{ji} = \psi_{jt}$), and each $\psi_{jt}$ is associated with one $\phi_l^k$, thus they form a shared structure across the corpus. Figure 1 visualize a sharing structure between first and second level DPs.

**Pólya urn scheme** A posterior perspective of the DP is the Pólya urn scheme which shows that draws from the DP are discrete and exhibit a clustering property. As Blackwell and MacQueen showed (Blackwell & MacQueen, 1973), our model can also be formed as a successive conditional distribution of $\theta_{ji}$ given $\theta_{j1}, ..., \theta_{ji-1}$.

Let $n_{jt}$ be the number of words for which factor $\theta_{ji}$ corresponds to $\psi_{jt}$ in document $j$, and $m_{jkl}$ be the number of $\psi_{jt}$ such that $\psi_{jt} = \phi_l^k$. Then the conditional distribution of $\theta_{ji}$ given $\theta_{j1}, ..., \theta_{ji-1}, G_0^1, ..., G_0^K$, and $\alpha$, with $G_j$ and $\lambda_j$ marginalized out, is

$$\theta_{ji}|\theta_{j1}, ..., \theta_{ji-1}, \alpha, \eta, G_0^1, ..., G_0^K \qquad (6)$$
$$\sim \sum_t \frac{n_{jt}}{i-1+\alpha} \delta_{\psi_{jt}} + \frac{\alpha}{i-1+\alpha} \sum_k \frac{m_{jk\cdot} + r_{jk}\eta}{m_{j\cdot\cdot} + |\boldsymbol{r}_j|\eta} G_0^k,$$

where $|\boldsymbol{r}_j|$ is the number of 1's in $\boldsymbol{r}_j$, and $r_{jk}$ is 1 if label $k$ has been observed in document $j$. $\theta_{ji}$ can be sampled from the first term of RHS or the second term of RHS. When $\theta_{ji}$ is sampled from the first term, then it corresponds to one existing $\psi_{jt}$, and when it is sampled from the second term, we choose $G_0^k$ to draw $\theta_{ji}$ with probability proportional to $m_{jk\cdot} + \eta$. After that, we can marginalize out $G_0^k$ to proceed further and get the conditional distribution

$$\psi_{jt}|\psi_{11}, ..., \psi_{jt-1}, \gamma_k, H_k$$
$$\sim \sum_k \frac{m_{\cdot kl}}{m_{\cdot k\cdot} + \gamma_k} \delta_{\phi_l^k} + \frac{\gamma_k}{m_{\cdot k\cdot} + \gamma_k} H_k. \qquad (7)$$

## 3. Inference via Gibbs Sampling

We propose a Gibbs sampler for DP-MRM, a Pólya urn scheme based on the marginalization of unknown dimensions (Escobar & West, 1995). For the collapsed Gibbs sampler, we marginalize out factors, $\theta, \psi, \phi$, mixing proportions, $\lambda$, and random probability measures, $G_j, G_0^k$. As a result, we only need to sample the index of each latent variable. Let $t_{ji}$ be the index variable such that $\psi_{jt} = \theta_{ji}$, and $k_{jt}$ be the index variable such that $\psi_{jt} \sim G_0^k$, and $l_{jt}$ be the index variable such that $\psi_{jt} = \phi_l^{k_{jt}}$. Let $n_{jt}$ be the number of $\theta_{ji}$ such that $\theta_{ji} = \psi_{jt}$, and let $m_{jkl}$ be the number of

Dirichlet Process with Mixed Random MeasuresTable 1. Datasets for single- and multi-labeled documents. The last two columns show the number of total labels in the corpus and the average number of labels per document.

|  | docs | vocab | labels | labels/doc |
|---|---|---|---|---|
| 20ng.comp | 1,800 | 2,608 | 5 | 1 |
| RCV1-V2 | 23,149 | 12,630 | 173 | 3.2 |
| Ohsumed | 7,505 | 7,056 | 52 | 5.2 |

$\psi_{jt}$ such that $\psi_{jt} = \phi_l^k$. We use $f_{kl}(x_{ji})$ to denote the conditional density of $x$ under mixture component $l$ of random measure $G_0^k$, given all items except $x_{ji}$,

$$f_{kl}(x_{ji}) = \frac{\int f(x_{ji}|\phi_l^k) \prod_{x_{j'i'} \in \mathbf{x}_{kl}} f(x_{j'i'}|\phi_l^k) h(\phi_l^k) d\phi_l^k}{\int \prod_{x_{j'i'} \in \mathbf{x}_{kl}} f(x_{j'i'}|\phi_l^k) h(\phi_l^k) d\phi_l^k},$$

where $\boldsymbol{x}_{kl} = \{x_{ji}; k_{jt_{ji}} = k, l_{jt_{ji}} = l\}$.

**Sampling $t$:** The conditional density of word $x_{ji}$ being assigned to $\psi_{jt}$ is

$$p(t_{ji} = t|t_{-ji}, rest)$$
$$= \begin{cases} \frac{n_{jt\cdot}}{n_{j\cdot\cdot}+\alpha} f_{k_{jt}l_{jt}}(x_{ji}) & \text{existing } t \\ \frac{\alpha}{n_{j\cdot\cdot}+\alpha} \Gamma(x_{ji}) & \text{new } t, \end{cases} \quad (8)$$

where $\Gamma(x_{ji}) = \sum_{k=1}^K \frac{m_{jk\cdot}+\eta}{m_{j\cdot\cdot}+K\eta} \sum_{l=1}^L \frac{m_{\cdot kl}}{m_{\cdot k\cdot}+\gamma_k} f_{kl}(x_{ji}) + \frac{\gamma_k}{m_{\cdot k\cdot}+\gamma_k} f_{kl_{new}}(x_{ji})$.

**Sampling $k$ and $l$:** When new $t$ is sampled, we need to sample $k_{jt_{new}}$ and $l_{jt_{new}}$. However, sampling $k$ and $l$ cannot be done independently because given $l$ the probability of $k$ is always zero except one. The joint conditional density of $k$ and $l$ is

$$p(k_{jt} = k, l_{jt} = l|\boldsymbol{k}_{-jt}, \boldsymbol{l}_{-jt}, rest) \quad (9)$$
$$\propto \frac{m_{jk\cdot}+\eta}{m_{j\cdot\cdot}+K\eta} \times \frac{m_{\cdot kl}}{m_{\cdot k\cdot}+\gamma_k} f_{kl}(x_{ji}) \quad \text{existing } l$$
$$p(k_{jt} = k, l_{jt} = l_{new}|\boldsymbol{k}_{-jt}, \boldsymbol{l}_{-jt}, rest) \quad (10)$$
$$\propto \frac{m_{jk\cdot}+\eta}{m_{j\cdot\cdot}+K\eta} \times \frac{\gamma_k}{m_{\cdot k\cdot}+\gamma_k} f_{kl_{new}}(x_{ji}) \quad \text{new } l.$$

Sampling $k$ and $l$ of existing $t$ changes the component memberships of all data items $\boldsymbol{x}_{jt} = \{x_{ji}; t_{ji} = t\}$, and this sampling can be done with the conditional distribution of $k$ and $l$ given $\boldsymbol{x}_{jt}$.

## 4. Application with Labeled Documents

We measure the performance of DP-MRM with three experiments. First, we compare the label prediction performance of DP-MRM and LDA-SVM on single-labeled documents. Then, we compare the label prediction performance of DP-MRM and L-LDA on multi-labeled documents. Finally, we compare the predictive performance of DP-MRM and L-LDA on heldout data. For the label prediction experiments, we take a semi-supervised approach: divide the corpus into training and test sets, infer the posterior distribution of the training set with the observed labels (i.e. $r_{jk} = 1$ only when $k \in label(j)$), and infer the posterior distribution of the test set with all possible $K$ labels (i.e. $r_{jk} = 1$ for all $k$).

For all evaluations, we run each model ten times with 5,000 iterations, the first 3,000 as burn-in and then using the samples thereafter with gaps of 100 iterations. For sampling the hyperparameters, we place Gamma(1,1) priors for $\gamma_k$, and $\alpha$, and set $\beta$ to 0.5.

### 4.1. Single-Labeled Documents

DP-MRM was designed to model multi-labeled documents, but it assumes that a label generates multiple topics, so this flexible assumption allows DP-MRM to be used for modeling single-labeled documents as well. Note that L-LDA for single-labeled documents would assign every word in a document to a single topic, and the document would thus be modeled as a mixture of unigrams (i.e., naive Bayes).

To measure the classification performance, we trained our model with five comp subcategories of newsgroup documents (20NG)[1]. Table 1 shows the details of our datasets. 90% of the documents were used with the labels, and the remaining 10% of documents were used without the labels. We classified each of the test documents by the label with the most number of words assigned. As a baseline, we trained a multi-class SVM with the topic proportions inferred by LDA (Blei et al., 2003). MedLDA (Zhu et al., 2009), one of supervised topic model, also used for the comparision. The results, shown in Figure 2, display a significant improvement of our model over the LDA-SVM approach and MedLDA.

### 4.2. Multi-Labeled Documents

We compared the performance of L-LDA and DP-MRM using two multi-labeled corpora: the Ohsumed dataset[2], which is a subset of the MEDLINE corpus consisting of medical journals, and RCV1-V2 dataset (Lewis et al., 2004), a corpus of Reuters news articles. We randomly sampled a subset of each corpus, and the detailed descriptions are shown in Table 1. Again, 90% of documents were used with the labels, and the rest 10% of documents were used without the labels.

L-LDA provides a systematic way of naming the dis-covered topics, and thus increases the interpretability of them. However, the assumption that a document is generated from a subset of topics specified by the observed labels limits the expressiveness of the model. DP-MRM was designed to keep the benefits of L-LDA while increasing the expressiveness, and we can see the consequences of the design in the discovered topics shown in Table 2. The table shows one label from each corpus and the corresponding topics. DP-MRM discovered multiple topics for the labels 'Infant' and 'Corporate/Industrial', and these are more detailed topics than the single topics discovered by L-LDA.

---
[1]http://people.csail.mit.edu/jrennie/20Newsgroups/
[2]http://ir.ohsu.edu/ohsumed/ohsumed.html

*Table 2.* Topics discovered by L-LDA and DP-MRM for the *Infant* label of the Ohsumed dataset, a corpus of medical journal articles, and for the *Corporate/Industrial* label of the RCV news articles corpus. We show the top ten probability words for each topic. L-LDA discovers exactly one topic per label, but DP-MRM discovers several topics per label.

| Infant | | | | Corporate/Industrial | | | | | |
|---|---|---|---|---|---|---|---|---|---|
| L-LDA | DP-MRM | | | L-LDA | DP-MRM | | | | |
| children | children | colon | tumor | compan | million | oil | shar | ton | airlin |
| infect | infect | aeruginosa | patient | million | profit | pow | compan | million | air |
| month | infant | express | leukemia | percent | percent | ga | bank | percent | carg |
| patient | month | gene | cell | market | half | compan | percent | produc | flight |
| ag | ag | type | chemotherapi | produc | expect | produc | million | export | servic |
| infant | antibodi | dna | dose | stat | compan | plant | invest | crop | airport |
| studi | hiv | mutat | therapi | bank | billion | operat | stock | wheat | carri |
| vaccin | vaccin | ha-ra | receiv | invest | result | refin | market | grain | plan |
| viru | viru | excret | treatment | plan | market | unit | stat | juli | operat |
| antibodi | test | urinari | remiss | billion | shar | million | plan | sugar | aircraft |

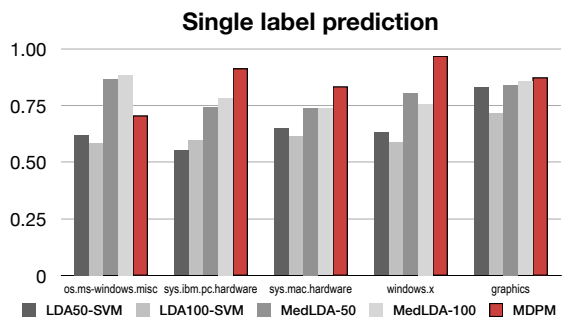

*Figure 2.* Accuracies of DP-MRM, MedLDA, and LDA-SVM on classification of 20NG. DP-MRM outperforms LDA-SVM and MedLDA on average.

For the classification of multi-labeled documents based on the posterior samples, we counted the number of words assigned to each measure $G_0^k$ and classified as label $k$ with various threshold cuts based on normalized counts. We scored each model based on Micro F1 and Macro F1 measures. Micro F1 accounts for the proportion of each class, so large classes affect its results, whereas macro F1 assigns equal weights to all classes. Table 3 shows the classification results with different cuts, and our model performs better than L-LDA in terms of micro average, but in macro average, there are inconsistencies between the different cuts. In general, DP-MRM shows more stable performance with respect to the cuts, whereas L-LDA shows variable results depending on the cut.

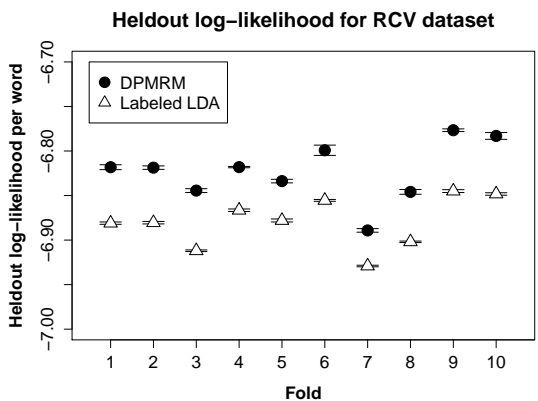

*Figure 3.* Heldout likelihood of DP-MRM and L-LDA for the RCV dataset. DP-MRM consistently outperforms L-LDA across the ten-folded heldout dataset.

### 4.3. Predictive Performance

To compare the model fit, we measure the predictive performance of our model and L-LDA with heldout likelihood of the test set. For each model, posterior sampling was done with 90% of the words in each document while the test set performance was evaluated on the remaining 10% of the words. Given $S$ samples from the posterior, the test set likelihood for our model is computed as follows:

$$p(\mathbf{x}^{\text{test}}) = \prod_{ji \in \mathbf{x}^{\text{test}}} \frac{1}{S} \sum_{s=1}^{S} \sum_{k=1}^{K} \sum_{l} \theta_{jkl}^{(s)} \psi_{klx_{ji}}^{(s)}$$

Table 3. Macro and micro F1 averages of L-LDA and DP-MRM for the two multi-label datasets. DP-MRM consistently performs better than L-LDA for micro F1, but not for macro F1.

|  | RCV | | | | Ohsumed | | | |
|---|---|---|---|---|---|---|---|---|
|  | Micro Average | | Macro Average | | Micro Average | | Macro Average | |
| Cut | DP-MRM | L-LDA | DP-MRM | L-LDA | DP-MRM | L-LDA | DP-MRM | L-LDA |
| 0.001 | 0.511 | 0.282 | 0.257 | 0.172 | **0.392** | 0.345 | 0.223 | 0.257 |
| 0.050 | 0.520 | 0.449 | 0.265 | 0.285 | 0.389 | **0.382** | **0.223** | **0.263** |
| 0.100 | **0.520** | **0.473** | 0.266 | 0.322 | 0.382 | 0.364 | 0.220 | 0.250 |
| 0.200 | 0.509 | 0.464 | 0.264 | **0.331** | 0.362 | 0.326 | 0.207 | 0.223 |
| 0.300 | 0.487 | 0.434 | 0.254 | 0.315 | 0.334 | 0.287 | 0.189 | 0.195 |
| 0.500 | 0.424 | 0.355 | 0.220 | 0.261 | 0.262 | 0.206 | 0.145 | 0.137 |

$$\theta_{jkl}^{(s)} = \frac{n_{jkl\cdot} + \alpha\{m_{\cdot kl}/(m_{\cdot k\cdot} + \gamma_k)\}}{n_{jk\cdot\cdot} + \alpha}$$

$$\psi_{klx_{ij}}^{(s)} = \frac{n_{\cdot klx_{ij}} + \beta}{n_{\cdot kl\cdot} + W\beta},$$

where $n_{jklx}$ is the number of words $x$ corresponding to $\phi_l^k$ in document $j$, and $W$ is the vocabulary size. The test set likelihood for L-LDA was computed as follows:

$$p(\mathbf{x}^{\text{test}}) = \prod_{ji \in \mathbf{x}^{\text{test}}} \frac{1}{S} \sum_{s=1}^{S} \sum_{k=1}^{K} \theta_{jk}^{(s)} \psi_{kx_{ji}}^{(s)}$$

$$\theta_{jk}^{(s)} = \frac{n_{jk\cdot} + \alpha}{n_{j\cdot\cdot} + K\alpha}$$

$$\psi_{kx_{ji}}^{(s)} = \frac{n_{\cdot kx_{ji}} + \beta}{n_{\cdot k\cdot} + W\beta},$$

where $K$ is the total number of labels. Figure 3 shows the test set per-word log likelihood of both model with RCV dataset, our model performs better than L-LDA across ten folded dataset consistently.

## 5. Image Segmentation with ddCRP

We describe an extension of DP-MRM, built by incorporating ddCRP, a nonparametric Bayesian prior that accounts for spatial dependencies, into DP-MRM. This illustrates the generality of DP-MRM that it may serve as a replacement for HDP for data with side information. We test this DP-MRM-ddCRP model on the task of image segmentation for multi-labeled images without manually segmented training data.

Image segmentation is often done with manually segmented and labeled data (He et al., 2004; Gould et al., 2009). DP-MRM can also perform supervised segmentation, but such data are harder to obtain, whereas image collections with multiple labels and no segmentation are relatively easy to obtain (e.g., Picasa or Flickr). One recent paper has shown a Bayesian model for simultaneous image segmentation and annotation (Du et al., 2009) using a logistic stick-breaking process. While that model is specialized for image understanding, DP-MRM is a general framework for modeling multi-labeled data including documents and images.

### 5.1. Incorporating ddCRP into DP-MRM

The Chinese restaurant process (CRP) is an alternative formulation of the DP. CRP forms a clustering structure of customers by assigning each customer to an existing or a new table. ddCRP, however, forms a clustering structure of customers by linking customers, accounting for the distances between them; customers who are relatively close to each other are likely to be linked together than those who are far apart. Let $c_i$ be the assignment of customer $i$ to the other customers, then the distribution of the customer assignment is

$$p(c_i = i' | c_{-i}, f, D, \alpha) \propto \begin{cases} f(d_{ii'}) & i \neq i' \\ \alpha & i = i' \end{cases}, \quad (11)$$

where $d_{ii'}$ is the distance between customer $i$ and $i'$, and $f(d_{ii'})$ is a decay function of the distance which mediates how the distances affect the resulting distribution over the partitions. There are many possible ways of defining the decay function, and in this paper, we follow (Ghosh et al., 2011) and use a *window* decay function which measures the distance between superpixels as a hop distance between them.

Based on the conditional distribution of assignments, the Pólya urn scheme for the combined model is:

$$\theta_{ji} | \theta_{j1}, ..., \theta_{ji-1}, \alpha, \eta, G_0^1, ..., G_0^K \qquad (12)$$

$$\sim \sum_{i'}^{i-1} \frac{f(d_{ii'})}{f_{sum}^i + \alpha} \delta_{\theta_{ji'}} + \frac{\alpha}{f_{sum}^i + \alpha} \sum_k \frac{m_{jk\cdot} + r_{jk}\eta}{m_{j\cdot\cdot} + |\mathbf{r}_j|\eta} G_0^k,$$

where $f_{sum}^i = \sum_{i' \neq i} f(d_{ii'})$. This equation is similar to Equation (6), but we modify the equation based on the window decay function.

For posterior inference, we modify the posterior sampling Equation (8) based on the customer assignment scheme, but the changes only affect the local sampling results (within the document level), and can be employed by the algorithm for ddCRP mixture in previ-ous work. The sampling scheme based on link structure among customers enhances the rapid mixing of sampler. See (Blei & Frazier, 2011) for a more detailed explanation of posterior inference.

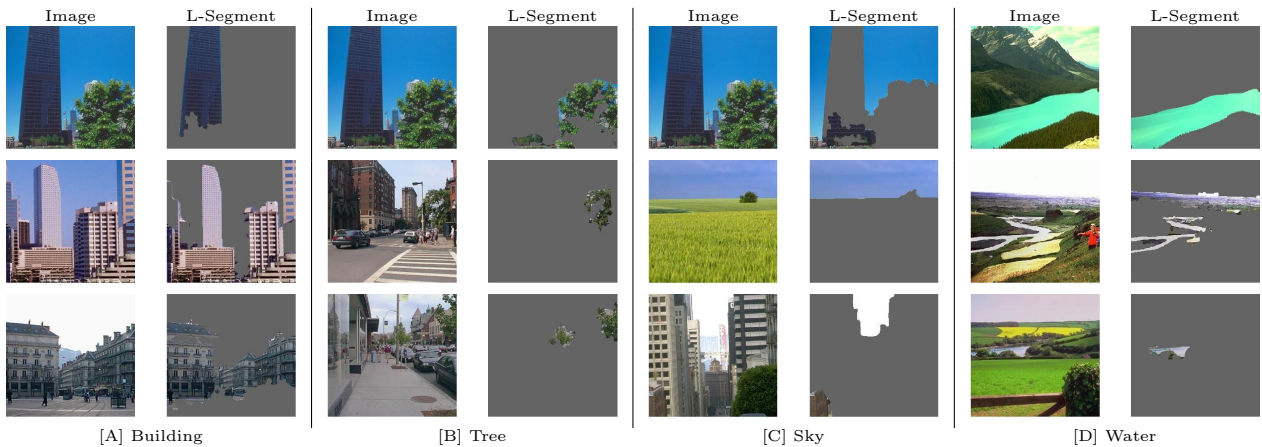

Figure 4. Labeled Segments (L-Segment) from posterior samples. From left to right, the segments are extracted from *building*, *water*, and *tree* labels. By using the list of objects without their location information, DP-MRM captures the image segments from co-occurrence patterns even when the object appears in a very small region (*water, tree*).

### 5.2. Image Segmentation with Multiple Labels

For image segmentation, we use the eight scene categories in (Oliva & Torralba, 2001) which are fully segmented and labeled by human subjects and available from the LabelMe dataset (Russell et al., 2008). A widely used method for representing images for inference is a codebook of images (Fei-Fei & Perona, 2005). To generate the codebook, each image is first divided into approximately 1,000 superpixels using the normalized cut algorithm (Shi & Malik, 2000). Each superpixel is described via local texton histogram (Martin et al., 2004) and HSV color histogram. By using $k$-means, we quantize these histograms into 128 bins, and superpixel $i$ in image $j$ is summarized via these codewords $x_{ji} = \{x_{ji}^t, x_{ji}^c\}$ indicating its texture $x_{ji}^t$ and color $x_{ji}^c$. The base distribution $H$ should be defined as $H \equiv \text{Dir}(\eta^t) \otimes \text{Dir}(\eta^c)$ for image segmentation.

Figure 4 shows some examples of the labeled objects from posterior samples where DP-MRM segments images into objects and labels each object. We note again that we do not give any pixel-level information for each object during the posterior inference, but our model can successfully segment images and label segments simultaneously. The results indicate that DP-MRM succeeds in inferring both the segments and the corresponding labels by capturing the co-occurrence patterns of superpixels and labels.

Figure 6 shows some examples of the image segmen-tation results comparing the original images, human segmented images, and DP-MRM segmented images. Figure 5 shows the quantitative performance of the segmentation via Rand Index, comparing DP-MRM with rddCRP (Ghosh et al., 2011) and normalized cuts (nCuts) (Shi & Malik, 2000), varying the number of segments from two to ten. We also vary the number of segments for each image, denoted as nCuts(*), where the number of segments are given as the number of labeled objects in each image. The result shows DP-MRM performs better than both rddCRP and nCuts.

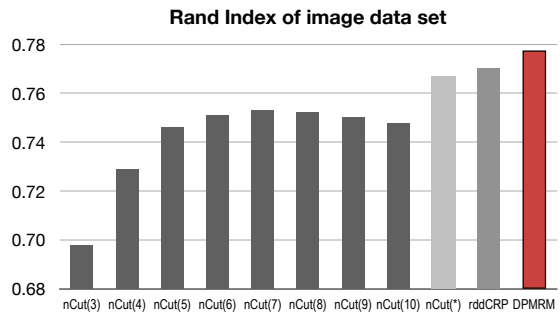

Figure 5. Rand Index of the segmentation results of nCuts, rddCRP, and DP-MRM on the LabelMe dataset. DP-MRM outperforms nCuts and rddCRP.

### 6. Conclusion

In this paper, we presented our new model, DP-MRM, in which the base distribution of DP is a mixture of random measures. The applications with multi-labeled documents and images are shown with label prediction and image segmentation experiments. The results show that DP-MRM for labeled data produces interpretable topics with more flexibility than the Labeled LDA. One promising extension of our model is to incorporate prior knowledge of external sources or domain experts into a Bayesian nonparametric topic model. It is beyond the scope of this paper, but our model can use different $\beta_k$ for each base distribution $H_k$, therefore using the structualized prior $\beta_k$ from domain experts (Andrzejewski et al., 2009) can be easily incorporated into our model.

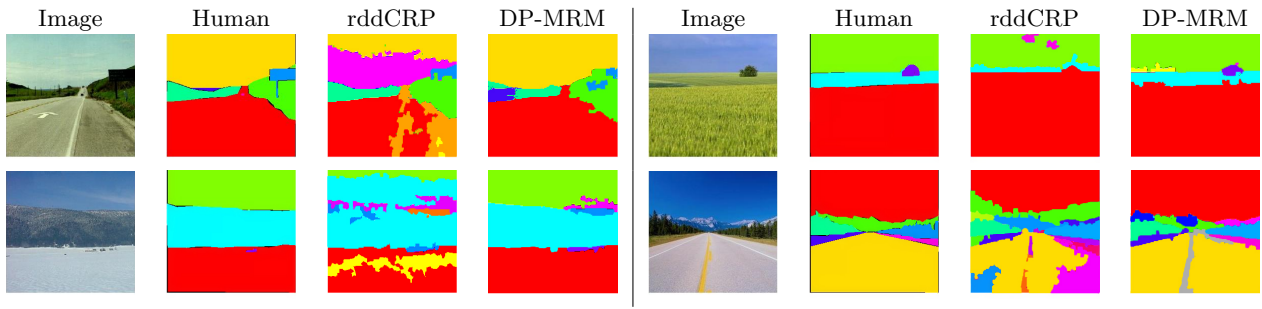

*Figure 6.* Segmentations produced by ddCRP-MRM. From left to right, the columns display natural image, segmentations by human, segmentation by rddCRP, and segmentations by ddCRP-MRM. Best viewed in color.

# Acknowledgments

This research was supported by Basic Science Research Program through the National Research Foundation of Korea (NRF) funded by the Ministry of Education, Science and Tehcnology (2011-0026507).